\documentclass{article}
\pdfoutput=1
\usepackage{hyperref,helvet,courier,subfig,dblfloatfix,setspace,ifthen,amsmath, amsthm, amssymb, graphicx,amsfonts,multirow,algorithm,algorithmic,times}

\newcommand*\Bell{\ensuremath{\boldsymbol\ell}}
\usepackage{mathrsfs,natbib}

\usepackage{natbib}

\usepackage{algorithm}
\usepackage{algorithmic}

\usepackage{hyperref}

\usepackage[accepted]{icml2016}

\icmltitlerunning{Generative Mixture of Networks}

\begin{document} 

\twocolumn[
\icmltitle{Generative Mixture of Networks}

\icmlauthor{Ershad Banijamali}{sbanijam@uwaterloo.ca}
\icmladdress{School of Computer Science, University of Waterloo}
\icmlauthor{Ali Ghodsi}{aghodsib@uwaterloo.ca}
\icmladdress{Department of Statistics and Actuarial Science, University of Waterloo}
\icmlauthor{Pascal Poupart}{ppoupart@uwaterloo.ca}
\icmladdress{School of Computer Science, University of Waterloo}
\icmlkeywords{Generative Model, Maximum Mean Discrepancy, Auto-encoder, Mixture Models}

\vskip 0.3in
]

\begin{abstract} 
A generative model based on training deep architectures is proposed. The model consists of $K$ networks that are trained together to learn the underlying distribution of a given data set. The process starts with dividing the input data into $K$ clusters  and feeding each of them into a separate network. After few iterations of training networks separately, we use an EM-like algorithm to train the networks together and update the clusters of the data. We call this model Mixture of Networks. The provided model is a platform that can be used for any deep structure and be trained by any conventional objective function for distribution modeling. As the components of the model are neural networks, it has high capability in characterizing complicated data distributions as well as clustering data. We apply the algorithm on MNIST hand-written digits and Yale face datasets. We also demonstrate the clustering ability of the model using some real-world and toy examples.
\end{abstract} 
\section{Introduction}
\label{Intro}
Deep architectures have shown excellent performance in various tasks of learning including, but not limited to, classification and regression, dimension reduction, object detection, and voice recognition.  In this work, we focus on another task, which is building a generative model. Generative models are used to characterize the underlying distribution of the data and then randomly generate samples according to their estimation of the distribution. Recently, use of deep architectures in the area of generative models is very popular among researchers. 

\subsection{Related Works}
A fundamental work on deep generative models has been done by Hinton et al. \yrcite{hinton2006}, where they introduced a fast algorithm for unsupervised training of deep belief networks (DBNs), which are deep graphical models. In a recent work by Salakhutdinov \yrcite{salakhutdinov2015learning}, a comprehensive review over this model is presented. Built upon this model, Lee et al. \yrcite{lee2009} presented a similar network with convolutional layers. They introduced probabilistic max-pooling technique and constructed a translation-invariant model. In \cite{ranzato2011}, another  generative model based on DBNs was presented, which was used for image feature extraction. Unsupervised deep representation learning techniques have been used in  \cite{bengio2013} to build a  generative model that can exploit high-level features to generate high-quality samples. 

Two important and recent classes of deep generative models are generative adversarial networks (GANs) \citep{goodfellow2014} and variational auto-encoders (VAE) \cite{vae2014}. 
GANs are trained based on solving a minimax problem to generate samples that are not distinguishable from the samples in  the training sets. Based on the variational inference concept,  VAEs are designed for fast training and having explicit expression for posterior probability of the latent variable. Many recent advancements in the area of deep generative models are based on these two models \citep{radford2015unsupervised,chen2016infogan,denton2015deep,sonderby2016ladder}.

Different types of neural networks have been used to work as a generative model for different applications. In \cite{gregor2015}, inspired by a human vision system, Recurrent Neural Networks (RNNs) are trained for generating images. \cite{dai2014} proposed a method for training Convolutional Neural Networks (CNNs) for this purpose.  In \cite{dosovitskiy2015}, authors trained a deep neural network in a supervised way to be able to generate images of different objects given their names, locations, and angles of view. 

In almost all of these works, the probability distribution of the output of the networks do not have a well-structured form. So, we have limited ability to extend these models and build a mixture model based on them. In mixture of expert models \cite{xu1995alternative,waterhouse1996bayesian}, we have networks as the component of a mixture of \textit{discreminative} models, where we assume some specific probability distribution on the output of the networks (i.e. Gaussian). However, such assumption for \textit{generative} models, where the output of the network is very high-dimensional, is not practical. 

\subsection{Contribution}
In this work, we introduce an algorithm for training mixture of generative models, which consists of deep networks as its components. Instead of using the whole training dataset to train one single network, our model is based on training multiple smaller networks by clusters of data. All the above-mentioned models can be components of this work to build a generative mixture model. The proposed method works under the assumption that components of a mixture model do not provide a closed form expression for the probability distribution of their output. We provide an algorithm which is inspired by expectation maximization (EM) to overcome this challenge. 

There are multiple advantages for the proposed algorithm compared to its predecessors, including:
\begin{itemize}
\item The accuracy of the output samples is higher, as each network is trained only with similar data points. 
\item After training the model, users can decide the \textit{category} of data they want to generate, instead of randomly generating samples.
\item The model, like other mixture models, can be used as a clustering method.
\end{itemize}

In the next section, the general idea of mixture models is shortly overviewed. Then, we describe the steps of the algorithm in detail. At last, the performance of the proposed algorithm is evaluated, both as a clustering and a generative model. 

\section{Background: Mixture Models and EM}
\label{sec: mixture model}
Mixture models are used to estimate the probability distribution of a given sample set where the overall distribution consists of several components. For the case of parametric mixture model, distribution of components are presumed to have some parametric form. Let $\theta$ denote the parameter set of a mixture model that has $K$ components, i.e. $\theta = \{\theta_1,\theta_2,...,\theta_K\}$, where $\theta_j$ represents the parameters of $j$th component. For example, if the components are assumed to be Gaussian, then $\theta_j = \{\mu_j,\Sigma_j\}$. A popular way to estimate $\theta_j$'s is using the Expectation Maximization (EM) algorithm. In the expectation (E) step of the EM algorithm, the membership probability of each data point is calculated for all components. In fact, it is the posterior probability over the mixture components. Let $m_{ij}$ be the probability of being a member of the $j$th component given the $i$th data point, $\mathbf{x}_i$, i.e.:
\begin{equation}
\label{eq: likelihood}
m_{ij} = P(j|\mathbf{x}_i,\theta_j)
\end{equation}
In the Maximization (M) step of the EM, the parameters of each component are updated using the membership probabilities. Each point contributes to updating the component parameters based on its component membership probability. To optimize the parameters, the E-step and M-step are consecutively taken until the algorithm converges to a local optimum or the maximum number of iterations is reached. 

\section{Mixture of Networks}
\label{sec: algorithm}
Inspired by the mixture models, we propose a combined generative and clustering algorithm. The learning process is completely unsupervised. Components of our model are constructed by networks. Neural networks have shown strong capabilities in estimating the distribution of complicated datasets. In fact, there is no constraint on the form of the distribution for components, e.g. Gaussian, Poisson, etc. Therefore, there is no parameter $\theta_j$ for the components to describe the distribution explicitly.  Instead, the parameters of our model are the weights in the networks that introduce an implicit probability distribution at the output of the networks. We denote the set of weights of the $j$th network by $\mathbf{w}_j$. 

We train multiple networks using the training data set. Similar to the EM algorithm, updating the network weights is based on membership probabilities. This means that we would like the points with high membership probability to an specific network to play significant role in further training that network. Therefore, the effect of different data points for updating the parameters of each network will be different. Each network tries to characterize one part of a multi-modal distribution function of the training data. In this section, we describe a mechanism that involves two steps of training and takes this problem into account.
\vspace{-.2cm}
\subsection{Steps of the training}
\textbf{Step 1:} The process starts with clustering the training set into $K$ partitions using a simple and hard-decision clustering algorithm, e.g. \textit{k}-means. By hard-decision, we mean the algorithm will assign data point $\mathbf{x}_i$ to exactly one cluster with probability one and to the rest of the clusters with probability zero. The knee method \cite{salvador2004} can be used to determine the number of clusters. Suppose $\mathit{X}$ is the training set with $N$ data points $X = \{\mathbf{x}_1,\mathbf{x}_2,...,\mathbf{x}_N\}$ in $\mathbb{R}^D$, and $\mathit{X}^j$ represents the $j$th cluster in this set, $\mathit{X} = \{\mathit{X}^1;\mathit{X}^2;...;\mathit{X}^K\}$. The clustering algorithm will divide the distribution of the training data into $K$ parts.  Each of these parts contains similar data points and has a smoother behavior compared to the original distribution of the whole data set.

Each cluster of the data is used to train one network, i.e. the  $j$th network is trained by the $j$th  cluster. Therefore, there will be $K$ networks. The structures of the networks, i.e. the number of layers and number of neurons in each layer, are identical. But, as we train networks with different subsets of the training set, parameters of the networks will be different. The ultimate goal for the networks is to minimize their defined cost function by adjusting their parameters. The cost function is a measure of dissimilarity between the training set and generated sample sets of a network. Let us denote the cost function for the $j$th network by $\mathcal{C}_j$. 
$\mathbf{w}_j^*$ is the optimum value for the $j$th network's parameters if and only if:
\begin{equation}
\mathbf{w}_j^* = \arg \min \limits_{\mathbf{w}_j} \mathcal{C}_j(X^j,Y(\mathbf{w}_j))
\end{equation}
where $Y(\mathbf{w}_j)$ is the set of samples generated by $j$th network. Mini-batch stochastic gradient descent (SGD) is used for training to find a local optimum for $\mathbf{w}_j$.
\begin{algorithm}[h]
   \caption{Training $j$th network by hard-decision clusters}
   \label{alg:initial training}
\begin{algorithmic}
   \STATE - Initialize the network parameter $\mathbf{w}_j$ randomly
   \FOR {$t=1$ {\bfseries to} $T_1$}
   \STATE - Divide $X^j$ randomly into $b$ mini batches of size $\mathcal{B}$. 
   \FOR{$i=1$ {\bfseries to} $b$}
   \STATE - Choose the $i$th mini batch of $X^j$  
   \STATE - Generate $\mathcal{B}$ output samples by the $j$th network using random input 
   \STATE - Update the network parameters
      \begin{equation}
   \label{eq: sgd}
   \mathbf{w}_j \gets \mathbf{w}_j - \alpha \frac{\partial \mathcal{C}_j}{\partial \mathbf{w}_j} 
   \end{equation}
   \ENDFOR
   \ENDFOR
\end{algorithmic}
\end{algorithm}

Algorithm \ref{alg:initial training}, describes the steps of the training of  $j$th network using the $j$th cluster.  The input of the network is a $p$-dimensional vector whose elements are drawn independently from a uniform distribution. The parameters of the networks are initialized randomly. Parameter $\alpha$ in (\ref{eq: sgd}) is the learning rate. The training process is done for $T_1$ epochs of each cluster. Let $\hat{\mathbf{w}}_j$ denote the parameters of network $j$ after this step of training. 

\textbf{Step 2:} After step 1, the output of the networks tend to be similar to their input datasets, which are clusters of the training set. Now, we propose an iterative model that works like the mixture models. It involves a process in which we further train the networks and cluster the training data set together. Clustering in this step is soft-decision, i.e. point $i$ belongs to cluster $j$ with membership probability $m_{ij} \in [0,1]$. Training the networks is also affected by these probabilities and different points will contribute differently in updating the parameters of the networks. However, instead of making any assumption on the distribution of the model's components, we propose an updating algorithm that is based on the output of trained networks in previous iterations. This means that if a data point is similar to the current outputs of one network, then it will have a high level of contribution in updating the parameters of that network in the next iteration.  Note that in step 2 of the algorithm, the whole training set is used to train each network.

To calculate the membership probabilities, we should have the probability distribution function for each component or network. As we did not impose such constraint on our model, we use kernel similarity between the data points and the generated samples of each network. In order to do this measurement, we generate $\mathcal{S}$ samples by each of the networks. $Y^j= Y(\hat{\mathbf{w}}_j)= \{\mathbf{y}_1^j,\mathbf{y}_2^j ,...,\mathbf{y}_{\mathcal{S}}^j \}$ represents the set of samples generated by $j$th network. Let $\ell_{ij}$ denote similarity of data point $\mathbf{x}_i$ to the samples in $Y^j$. Then:
\begin{equation}
\ell_{ij} = p(\mathbf{x}_i|\hat{\mathbf{w}}_j) = \frac{1}{\mathcal{S}} \sum \limits_{r=1}^{\mathcal{S}} k(\mathbf{x}_i,\mathbf{y}_r^j)
\end{equation}
The kernel that we use here is Gaussian.

The membership probability also needs the prior probability over each component, which is denoted by $\pi_j$. The initial value of $\pi_j$ in step 2 is: $\pi_j = |X^j|/N$. Here, the membership probability is interpreted as the probability that network $j$ has produced data point $\mathbf{x}_i$, and is given by:
\begin{equation}
m_{ij} = \frac{\ell_{ij} \pi_j}{\sum_{k=1}^K \ell_{ik} \pi_k}
\end{equation}
Note that we should have $\sum_{r=1}^K m_{ir} = 1$. Value of the prior probabilities after the first iteration in this step is updated by: $\pi_j = (\sum_{i=1}^N m_{ij})/N$. Similar to the EM algorithm, we want the effect of point $\mathbf{x}_i$ in updating parameters of network $j$  ($\hat{\mathbf{w}}_j$) to be proportional to $m_{ij}$. To do so, we multiply the membership probabilities to the learning rate of the SGD algorithm. If a membership is high, then the learning rate will be high and the effect of that point will be high. If the membership probability is low and near zero, the learning rate will be near zero and the algorithm will not update the network parameters based on that point. 

Lets call $\Bell_i = \{\ell_{i1},\ell_{i2},...,\ell_{iK}\}$ the likelihood vector assigned to the $i$th data point. Suppose that likelihood of all points for all networks are stored in an $N \times K$ matrix $\mathscr{L}$. Each row of this matrix is corresponding to one point in the training set. Using the above procedure, this matrix is updated iteratively (after using each epoch to update all networks parameters). The initial value of likelihood matrix is obtained by generating $\mathcal{S}$ samples using each of the trained networks in the step 1. 

In order to accelerate the learning process, we use mini-batch SGD here, as well.  We need to define mini-batch membership. The membership of mini-batch $b_i$ of size $\mathcal{B}$ for the $j$th network is defined as $m_{b_i,j} = P(j|\hat{\mathbf{w}}_j,\{\mathbf{x}_r \in b_i\})$ and:
\begin{equation}
\label{eq: batch likelihood}
m_{b_i,j} = \cfrac{p(\{\mathbf{x}_r \in b_i\}|\hat{\mathbf{w}}_j) \pi_j}{\sum\limits_{k=1}^K p(\{\mathbf{x}_r \in b_i\}|\hat{\mathbf{w}}_k)\pi_k} = \cfrac{\pi_j \prod \limits_{\mathbf{x}_r \in b_i} \ell_{rj}}{\sum \limits_{k=1}^K \pi_k\prod \limits_{\mathbf{x}_r \in b_i} \ell_{rk}}
\end{equation}
For training the network $j$ using $b_i$, the learning rate is multiplied to $m_{b_i,j}$. According to (\ref{eq: batch likelihood}), $m_{b_i,j}$ contains the effect of $\mathcal{B}$ points together. However, these $\mathcal{B}$ points do not necessarily have similar likelihood vectors. So, the multiplication in (\ref{eq: batch likelihood}) can mix the effect of important and non-important points for training an specific network. To solve this issue we should somehow put points which are important for training a network together. A systematic solution is to rearrange the rows of the  likelihood matrix $\mathscr{L}$ at each iteration of the step 2, such that the first $N_1$ rows have the maximum likelihood in their first columns, the next $N_2$ rows have the maximum likelihood in their second columns, and so on. Where $N_j = |\{\mathbf{x}_i|\ell_{ij} \geq \ell_{ik} \text{ , } \forall k\neq j \}|$ and obviously $\sum_{k=1}^K N_k = N$. The process is similar to the bootstrap sampling. The corresponding data points to the rows of $\mathscr{L}$ are also rearranged in the same way. 
\begin{algorithm}[!h]
   \caption{Training networks using soft-decision clusters}
   \label{alg:final training}
\begin{algorithmic}
   \STATE  - Initialize likelihood matrix $\mathscr{L}$ based on the clusters in Step 1
   \FOR {$t=1$ {\bfseries to} $T_2$}
   \STATE  - Rearrange data set $X$
   \STATE  - Divide $X$ into $\lfloor \frac{N}{\mathcal{B}} \rfloor$ mini-batches of size $\mathcal{B}$. 
   \STATE  - Compute the mini-batch memberships. 
   \FOR {$j=1$ {\bfseries to} $K$}
   \STATE Choose $j$th network	
   \FOR{$i=1$ {\bfseries to} $\lfloor \frac{N}{\mathcal{B}} \rfloor$}
   \STATE - Choose $i$th mini-batch of $X$  
   \STATE - Generate $\mathcal{B}$ samples by $j$th network
   \STATE - Update the network parameters
   \vspace{-.2cm}
   \begin{equation}
   \vspace{-.2cm}
   \label{eq: sgd_2}
   \hat{\mathbf{w}}_j \gets \hat{\mathbf{w}}_j - m_{b_i,j} \times \beta \frac{\partial \mathcal{C}_j}{\partial \hat{\mathbf{w}}_j} 
   \end{equation}
   \ENDFOR
   \ENDFOR
   \STATE - Update the likelihood matrix $\mathscr{L}$ 
   \ENDFOR
\end{algorithmic}
\end{algorithm}

Algorithm \ref{alg:final training} summarizes the described procedure in the step 2. Rearranging data set $X$ in this algorithm refers to the procedure stated above. Note that dividing the data into mini-batches is not random in the step 2.   
The whole process of this step is done for $T_2$ epochs or iterations.

After this step, the training process is finished. Now we have a hyper-network consists of $K$ small networks with similar structures but different parameters. To generate samples randomly using the hyper-network, one of the networks is randomly chosen based on the priors. That is, the $j$th network is chosen by probability $\pi_j$. To generate a sample from a specific cluster, the corresponding network should be picked manually. Then using a random input, the selected network generates the desired sample.

A feature that distinguishes this model from the previous unsupervised generative models is its capability to generate a specific type of sample. For example, if the networks are trained over a set of face images with different expressions, then it can be used to generate a face in a special category (age, expression, illumination, and etc.), e.g. "a laughing old man", instead of generating samples randomly and waiting for our desired output. This can have many applications including automatic visualization of text.

\subsection{Maximum Mean Discrepancy as the cost function}
The proposed structure in this paper can be trained by any conventional objective function at the output (for example the objective in \cite{dosovitskiy2015}). However, here we use the maximum mean discrepancy ($\textit{MMD}$), introduced by Gretton et al. \yrcite{gretton2006kernel}, because of its simplicity and effectiveness.  $\textit{MMD}$ was also used in two recent works \cite{dziugaite2015,li2015}.  Therefore, the model parameters are learned based on minimizing the distance between the distribution of the samples generated by the network and samples from the training set, using $\textit{MMD}$.

Suppose $\mathbf{x}$ has distribution $p$ and $\mathbf{y}$ has distribution $q$. Let $\mathcal{F}$ be a class of functions. The squared population $\textit{MMD}$ is:
\begin{equation}
\textit{MMD}^2(\mathcal{F},p,q) = \big[ \sup_{f \in \mathcal{F}} \big( \mathbf{E}_x[f(\mathbf{x})] - \mathbf{E}_y[f(\mathbf{y})] \big)\big]^2.
\end{equation}

If $\mathcal{F}$ is a class of functions in the unit ball in a universal Reproducing Kernel Hilbert-Space (RKHS), then $\textit{MMD}$ is zero if and only if $p=q$ \cite{gretton2012kernel}. In this case, $\textit{MMD}$ can also be written in the form of a continuous kernel in that RKHS. 

However, in our applications, the underlying \textit{pdf} of the sample sets are unknown.  Suppose we have two sample sets $\mathit{X} = \{\mathbf{x}_1,\mathbf{x}_2,...,\mathbf{x}_M \}$ and $\mathit{Y} = \{\mathbf{y}_1,\mathbf{y}_2,...,\mathbf{y}_N \}$. The unbiased empirical estimation of the squared $\textit{MMD}$, according to \cite{gretton2012kernel}, for these two sets is given as: 
\begin{equation}
\label{eq: empirical mmd}
\begin{array}{c}
\textit{MMD}^2(\mathcal{F},\mathit{X},\mathit{Y})\!=\!\frac{1}{M(M-1)}\sum \limits_{i=1}^M \sum \limits_{j\neq i}^M k(\mathbf{x}_i,\mathbf{x}_j) \\
 + \frac{1}{N(N-1)}\sum \limits_{i=1}^N \sum \limits_{j\neq i}^N k(\mathbf{y}_i,\mathbf{y}_j) \!-\! \frac{2}{MN}\sum \limits_{i=1}^M \sum \limits_{j=1}^N k(\mathbf{x}_i,\mathbf{y}_j)
\end{array}
\end{equation}
We will use Gaussian kernel here too.

\subsection{Making the Algorithm Faster and More Effective}
Our results show that the batch membership can be very small for most of the batches. So, for each step of training of the networks, we only use the batches that have membership probability more than a threshold (in our experiments $0.001$) and do not use the rest of the batches that have negligible batch memberships. This way, the training process will be much faster.

In \cite{ramdas2015}, authors have shown that the power of kernel-based methods, such as $\textit{MMD}$,  for two-sample test problem drops polynomially with increasing dimensions. This suggests that a dimension reduction is helpful as a data pre-processing step for high-dimension datasets. Using an autoencoder is a solution here. We train an autoencoder separately using the complete training dataset. The networks in this scenario should be trained using a low-dimensional version of data. At the output of the generative networks, the decoder part of the autoencoder is used to map the data back into the original space. The hard-decision clustering in the first step of our algorithm can be either performed on the original data or its low-dimensional version.

\section{Experiment Results}
\label{sec: results}
To highlight the clustering capability of the proposed algorithm, we first apply it on synthetic toy datasets and real-world datasets. Then, we apply the algorithm on two real-world datasets: MNIST hand-written digits dataset and the Yale Face Database. In all of these experiments, the components of the model are fully-connected networks with multiple layers. Input to the networks is a random vector with elements drawn independently from uniform distribution in $[-1,1]$. The number of layers, number of hidden units in each layer, and the dimension of random input depend on the dataset.  The activation function for all hidden layers is ReLU and sigmoid for the output layer. All hyper parameters of the model are set by validation. For each dataset, we keep a portion of data points only for validation. This portion is not used for training. The validation set is also used to prevent overfitting. We continue the training until the average log-likelihood of the validation set is saturated.

\begin{table*}[!b]
\vspace{0cm}
\caption{\small Comparison of clustering purity ($\%$) for different datasets. The bold numbers show the best results among these algorithms. $d=$ dimensionality of the original space, $n=$ dataset size, $p=$ dimensionality of the low-dimensional dataset using autoencoder, $L$ = \# of classes}
\begin{center}
\small
\begin{tabular}{lcccccccccc}
\hline\noalign{\smallskip}
\small
Dataset & $d$ & $n$ & $p$ & $L$ & $k$-means & NCut   & LLC & LDMGI & Mix. of Nets & Networks Structure
\\
\noalign{\smallskip}
\hline
\hline
COIL-20		&	1024 & 1440		& 32  	& 20	& 62.3$\pm$3.1 & 68.4$\pm$5.3  & 67.5$\pm$5.1  & 75.3$\pm$4.9  & \textbf{77.6}$\pm$\textbf{3.1}  & 10-16-256-256-512-32 \\
\hline
Reuters-10K	&	2000 & 10000		& 128  	& 4		& 53.1$\pm$2.8 & 59.3$\pm$4.2  & 57.1$\pm$3.9  & 43.2$\pm$3.7  & \textbf{63.1}$\pm$\textbf{4.2}  & 12-64-256-512-512-128 \\
\hline
USPS		&	256 & 9298		& 32  	& 10	& 64.9$\pm$3.6 & 73.4$\pm$6.3  & 70.1$\pm$3.9  & \textbf{80.5}$\pm$\textbf{5.6}  & 78.3$\pm$3.7  & 10-16-256-256-512-32 \\
\hline
Isolet	&	7797 & 617	& 32  	& 26	& 63.7$\pm$2.8 & 65.7$\pm$3.4  & 69.3$\pm$2.7  & 68.8$\pm$3.6 & \textbf{71.3}$\pm$\textbf{3.0}  & 12-32-256-256-512-32 \\
\hline
\end{tabular}
\end{center}
\label{tbl: comp_CP}
\vspace{0cm}
\end{table*} 

\subsection{Performance as a clustering algorithm}
\subsubsection{Toy datasets}
In this section, we use three small toy datasets to visualize the clustering performance of the algorithm. We call these datasets two-moon, moon-circle, and two-circle. The first two datasets have $4000$ data points and the last one has $4500$ data points. All datasets have two dimensions. The datasets are first divided into two parts using $k$-means and then fed to the model. The model has two networks. We used similar structure for the networks for all datasets. The networks has $3$ hidden layers with $32$, $128$, and $32$ hidden units. The input to the network is 2-dimensional. For all of the experiments $T_1 = 30$, $T_2 = 200$, and $\mathcal{B} = 100$.

\begin{figure}[!t]
\captionsetup[subfigure]{labelformat=empty}
    \centering
    \subfloat[(a)]{{\includegraphics[trim = 40mm 80mm 30mm 80mm,height=3cm]{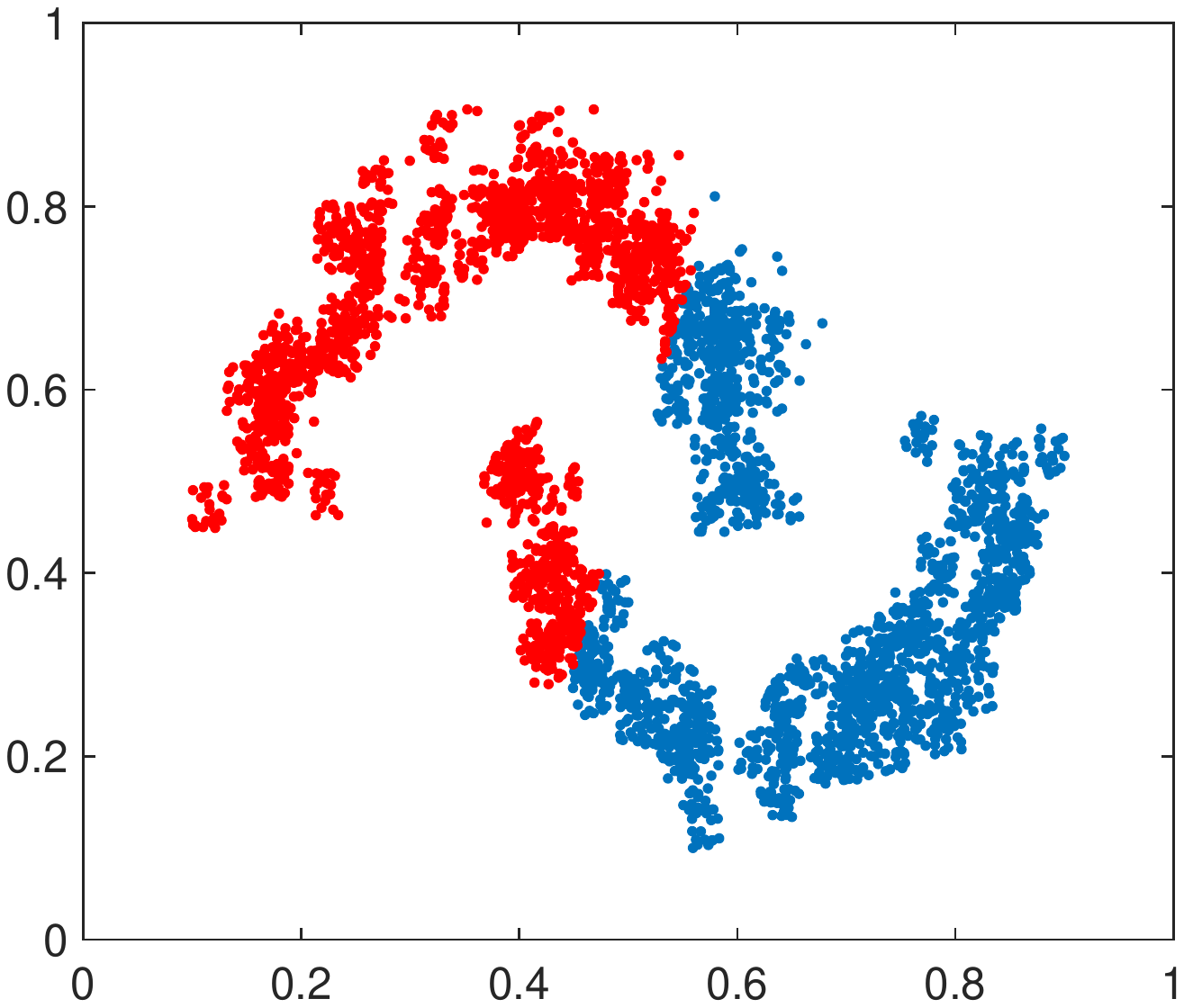} } 
    {\includegraphics[trim = 50mm 80mm 30mm 80mm,height=3cm]{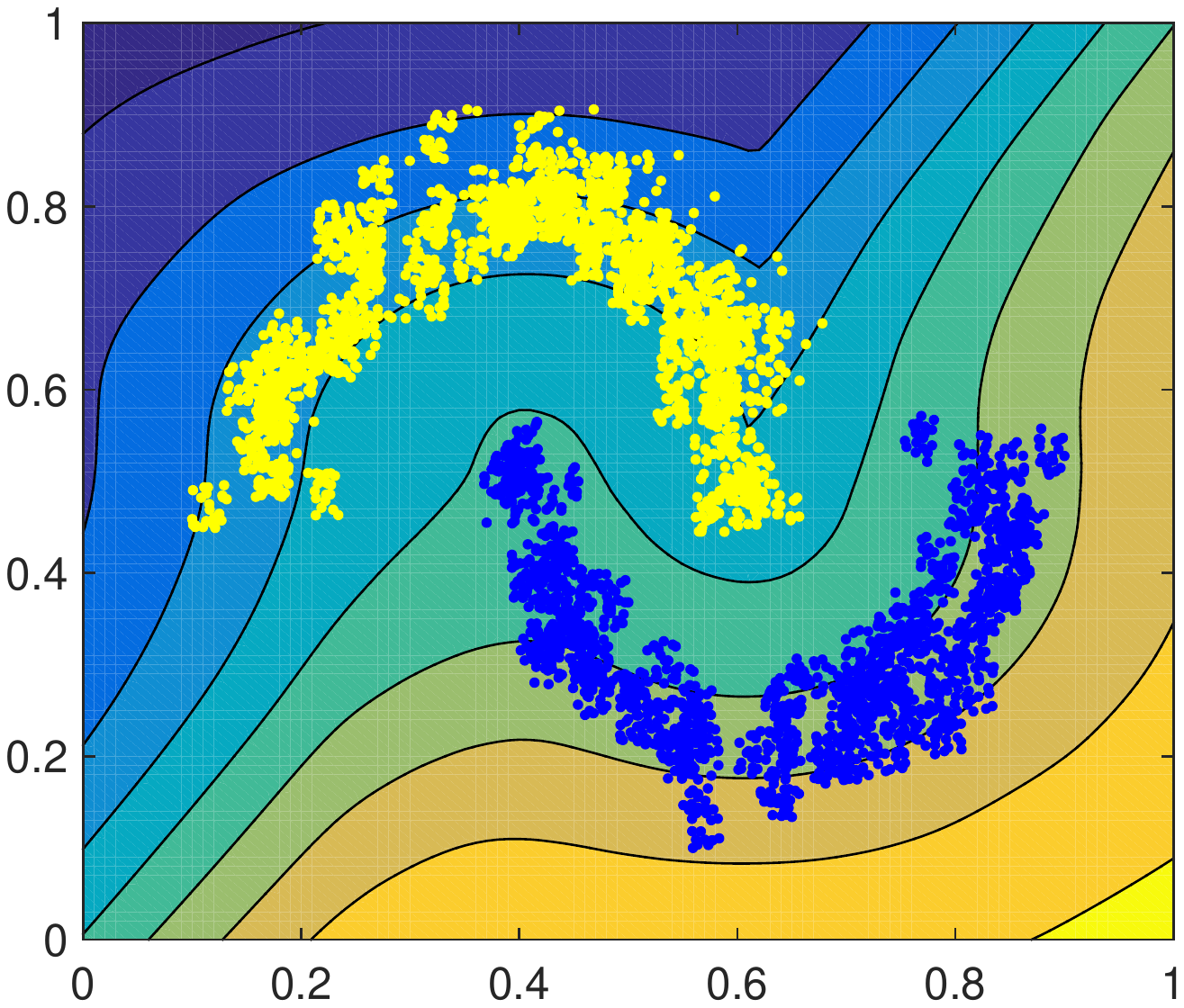} }} \\
	\subfloat[(b)]{{\includegraphics[trim = 40mm 80mm 30mm 80mm,height=3cm]{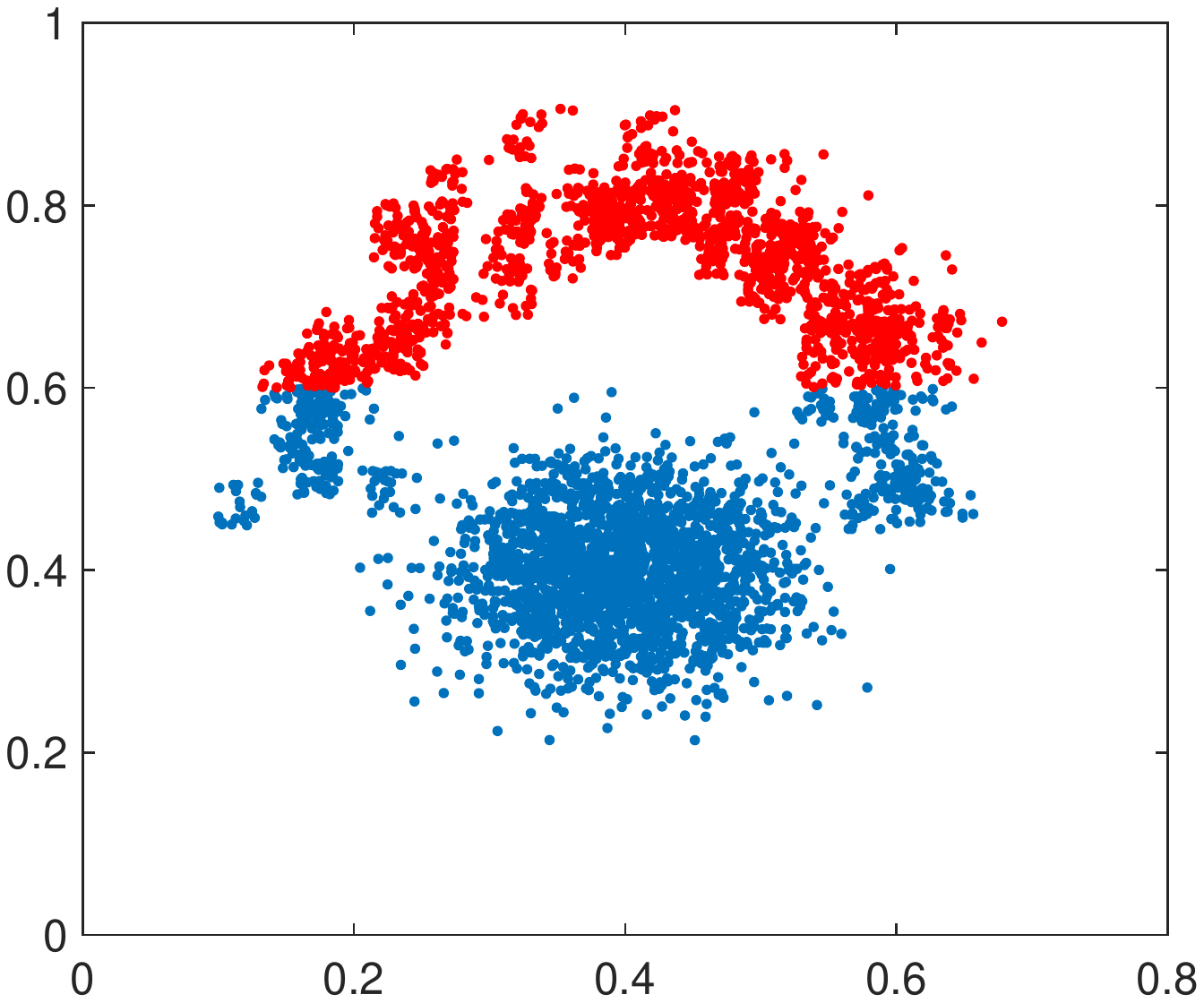} }
	{\includegraphics[trim =50mm 80mm 30mm 80mm,height=3cm]{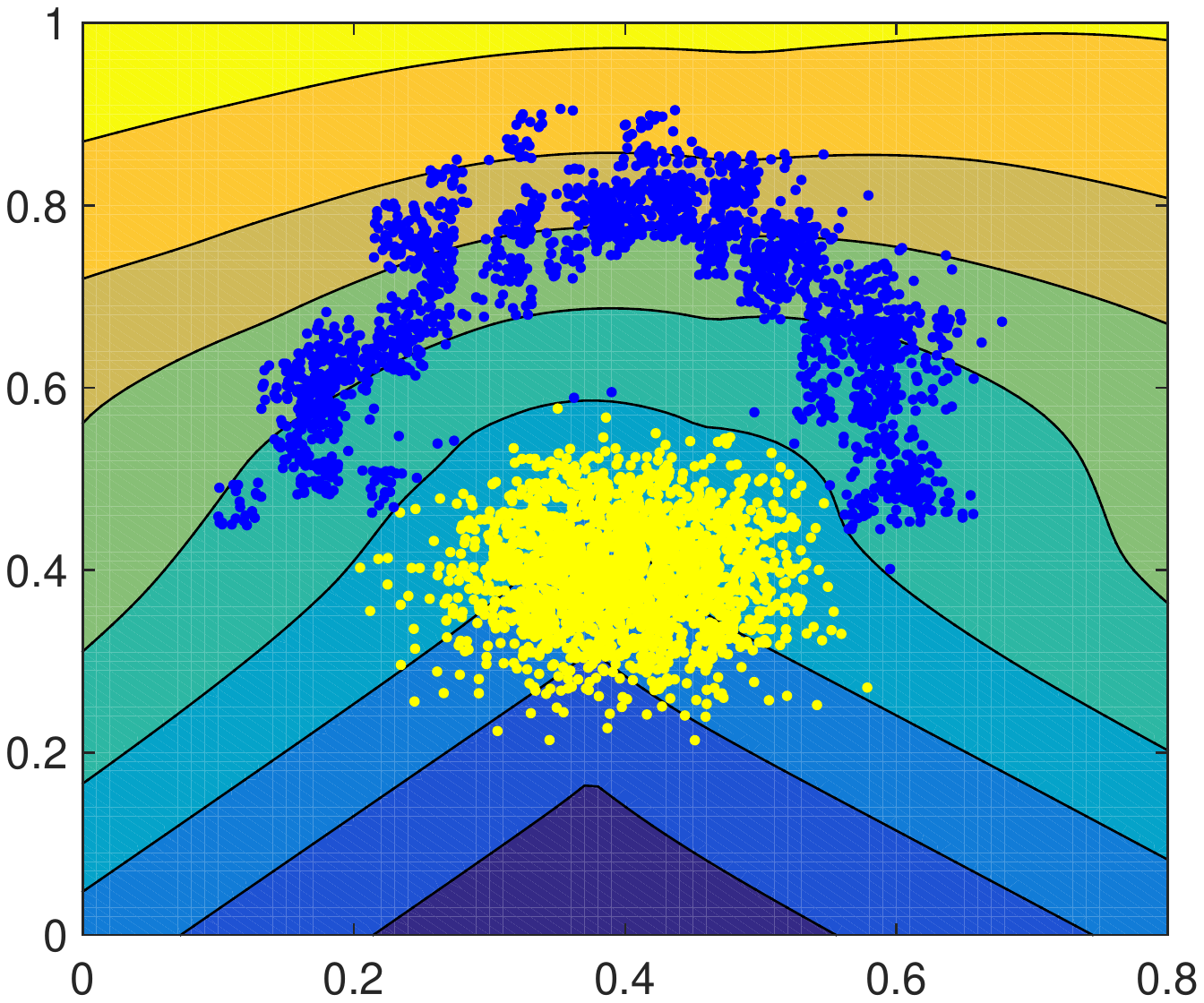} }} \\
	\subfloat[(c)]{{\includegraphics[trim = 40mm 80mm 30mm 80mm,height=3cm]{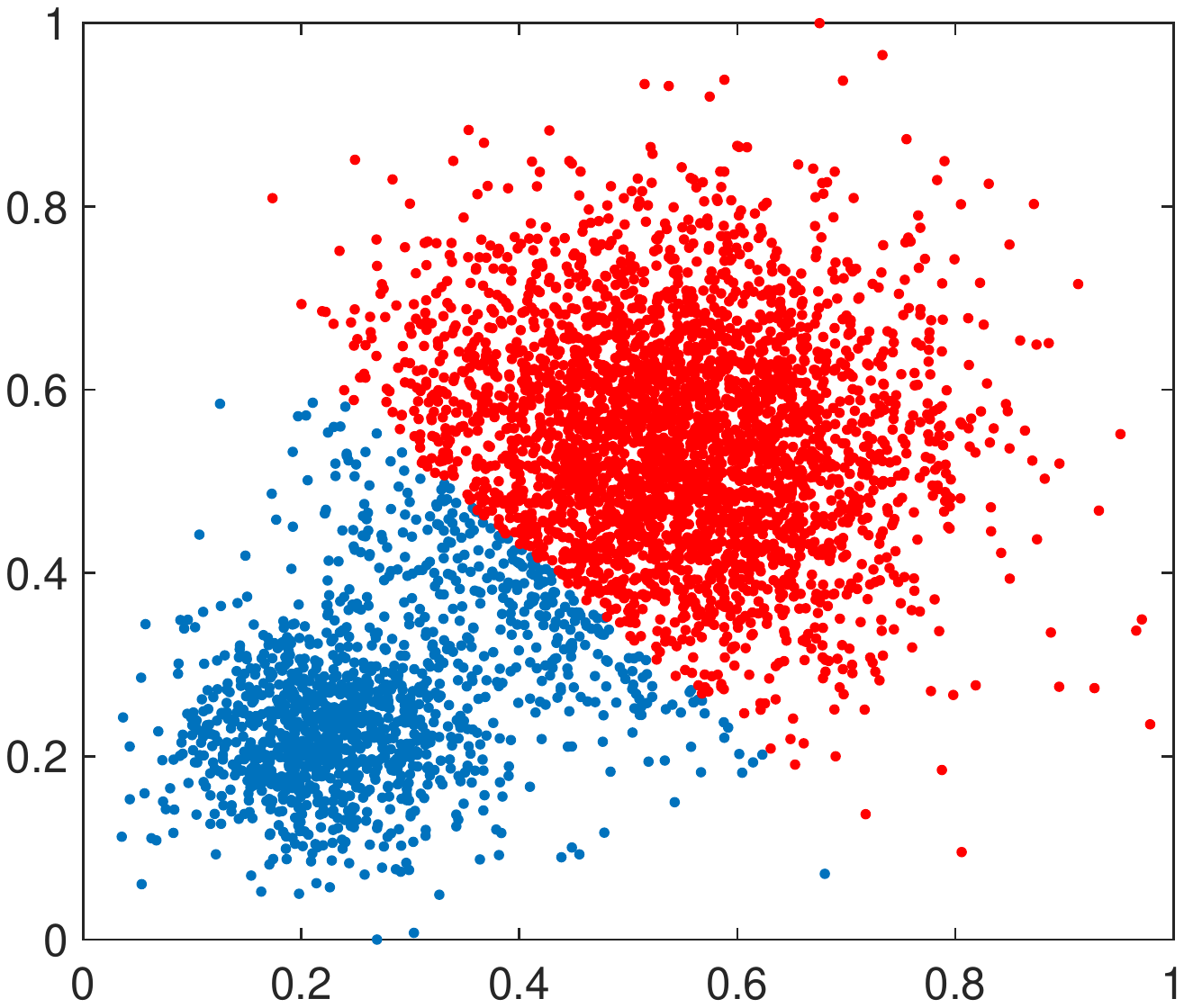} }
	{\includegraphics[trim =50mm 80mm 30mm 80mm,height=3cm]{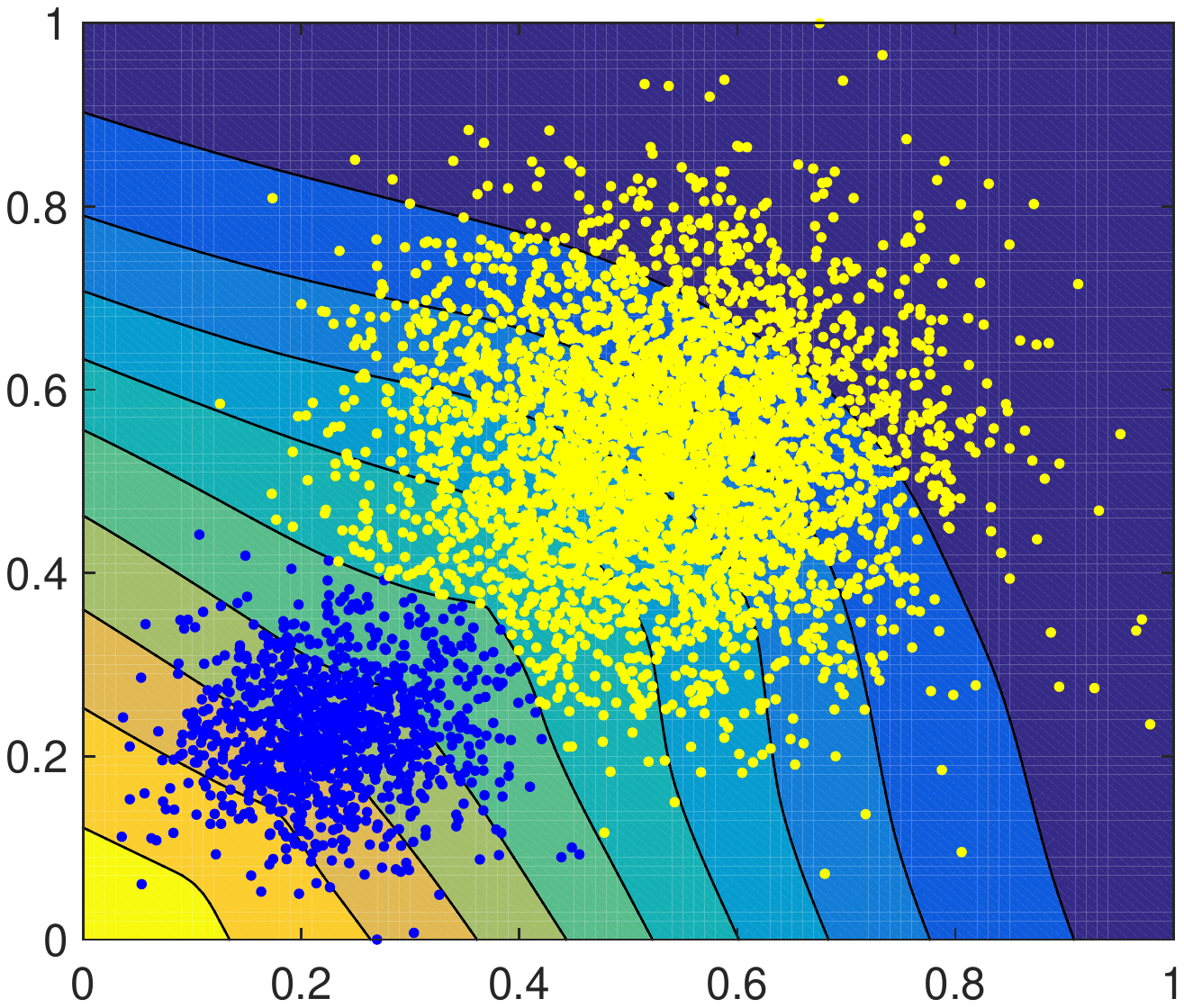} }}
   
     \small   
    \caption{(a), (b), and (c) represent three different datasets. The left figure shows the initial clustering by $k$-means and the right figure shows the final clusters using the proposed method as well as the contour of membership probability and its isolines.}
    \label{fig: toy}%
\end{figure}

Fig. \ref{fig: toy} shows the results of these experiments. As we can see, the algorithm could learn the model parameters to identify the natural clusters. This shows that the algorithm can successfully characterize the distribution of data clusters. Conventional mixture models, such as Gaussian mixture model, obviously fail to identify these clusters for two-moon and moon-circle. Clustering algorithms based on similarity matrices, such as spectral clustering, could have also identified clusters, but they do not possess the generative aspect of our model. Besides, these algorithms usually include an eigen-decomposition step, which is very computationally expensive when it comes to clustering large datasets.

\subsubsection{Real-world datasets}
Here, we also evaluate the clustering performance of the algorithm for some real-world datasets based on \textit{clustering purity (CP)}. \textit{CP} is defined for a labeled dataset as a measure of matching between classes and clusters. If $\{C^1,C^2,...,C^L\}$ are $L$ classes of a dataset $X$ of size $N$, then a clustering algorithm, $\mathcal{A}$, which divides $X$ into $K$ clusters $\{X^1,X^2,...,X^K\}$ has $\textit{CP}(\mathcal{A},X)$ as:
\vspace{-.15cm}
\begin{equation}
\textit{CP}(\mathcal{A},X) = \frac{1}{N} \sum \limits_{j=1}^K \max \limits_i |C^i \cap X^j|. 
\end{equation}
Note that we specify the final clusters by assigning each point to the cluster with highest membership probability. 

Table \ref{tbl: comp_CP} shows the results of clustering for four different datasets. 1) COIL-20: $32\times 32$ images of $20$ different objects from different angels. Dataset has $72$ images in each class. 2) Reuters-10K: Reuters dataset \cite{Lewis2004}, contains $810000$ English news stories in different categories. We followed the same procedure in \cite{xie2016} to obtain $10000$ samples from this set in $4$ categories. 3) USPS: This dataset contains $16\times 16$ images of hand-written digits. 4) Isolet: This is from UCI repository and contains the spoken alphabet letter from different individuals. Other statistics of the datasets are mentioned in the table. Number of clusters for these experiments are chosen to be equal to number of classes.

We compared the performance of the algorithm with $4$ other algorithms. $k$-means, which is used as the initial clustering for our method as well. The other three algorithms are based on spectral clustering. NCut is the classic spectral clustering, which assigns cluster labels to the data points by running $k$-means on the eigenvectors of the Laplacian matrix of dataset graph. Local linear approach for data clustering (LLC) \cite{wu2006}, assigns cluster labels to each data point based on linear combination of the kernel similarity between that point and its neighbors. Finally, local discriminant models  and global integration (LDMGI) \cite{yang2010}, which introduces a novel method for the learning Laplacian matrix by employing manifold structure and local discriminant information. LDMGI is designed specially for image clustering. We run experiments $10$ times to obtain the results in the table. As we can see, the proposed algorithm achieves the best or near-best results for different datasets.

\begin{figure*}[!t]
\vspace{0cm} 
    \centering
    \includegraphics[trim = 00mm 0mm 00mm 0mm,width=17.5cm]{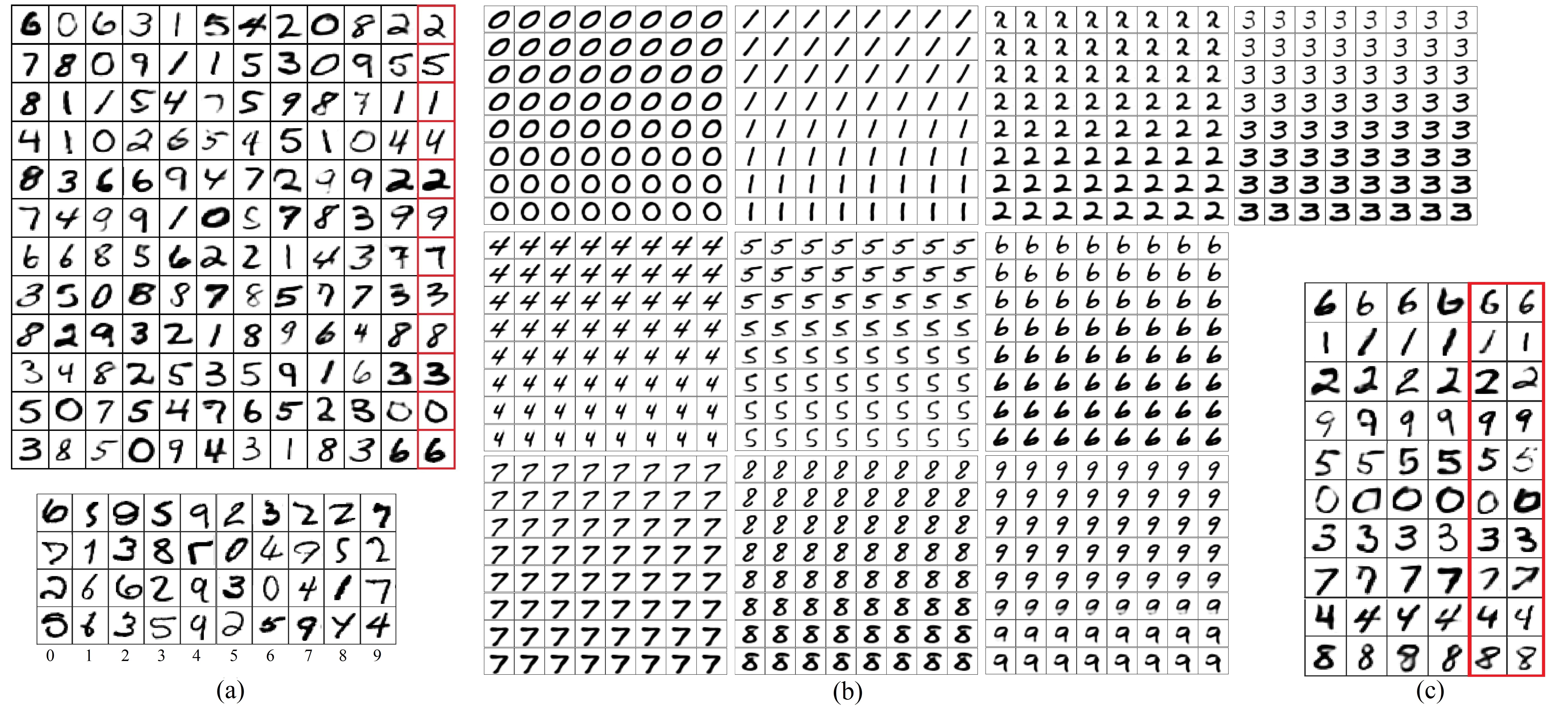}     
\caption{(a) Top: Samples generated randomly by the hyper-network using the cluster priors. The right most column shows samples from the training set which are nearest-neighbors of their adjacent images in the low-dimension space. This column is added to show that the generated samples are not merely a copy of the training samples. Bottom: Examples of digits of different classes that are mis-clustered by $k$-means and NCut but mixture of networks clustered them correctly. (b) For each of these 10 sub-images only one networks has been chosen to generate data. For each sub-image we generate two data points with different shapes (top-left corner and bottom-right corner) using two different random inputs. By traversing on an straight line in the latent space we obtain the other data points in the sub-image. As we can see, this shows that the model learns a proper mapping between the latent space and the data space. (c) Digits is each row are generated using one network. Digits in each of the last two columns are generated by giving identical input to different networks.}  
\label{fig: mnist}
\vspace{-.4cm}
\end{figure*}

\subsection{Performance as a generative model}
\subsubsection{MNIST dataset}
Samples of MNIST set are $28\times 28$ images of hand-written digits. The dataset consists of $60000$ training samples and $10000$ test samples. We use $5000$ samples in the training set for validation and the rest for training networks.

We first train an autoencoder, which maps the original data to a $32$-dimension space. Although, the knee method suggested $12$ clusters here, we divided the training set into $10$ clusters using \textit{k}-means to see if we can capture each class by a single network. So, we will have $10$ networks to be trained. The networks have $4$ hidden layers with $64$, $256$, $256$, and $512$ units. Input of the networks is $12$-dimensional. Each network is first trained by Algorithm \ref{alg:initial training} using its corresponding cluster for $30$ epochs ($T_1=30$).  Then using Algorithm \ref{alg:final training}, the data points membership probabilities and network parameters are updated up to $T_2=200$ iterations. Batch size for both steps is $100$. We use weight decay as regularization to improve the generalization of the model.

We repeated the whole process of training ten times. Using \textit{k}-means, the initial value of \textit{CP} on the low dimensional version of the data is $59.2 \pm 3.1$. After applying our metho,d \textit{CP} goes up to $80.3 \pm 4.2$, which is close to the state-of-the-art clustering methods on MNIST according to \cite{xie2016}. 

Fig. \ref{fig: mnist}, shows the samples generated by our model. An evaluation measure that is commonly used for generative models is the average log-likelihood of the test set, also known as Parzen estimation. We generated $10000$ samples randomly by the model and fit a Gaussian Parzen Window. We report our model's average log-likelihood of the test set for MNIST as $\textbf{308} \pm \textbf{2.8}$. Table \ref{mnist table}, shows a comparison between different methods in terms of average log-likelihood. However, based on \cite{theis2015}, this evaluation for generative models can be misleading, as samples generated by a naive methods may achieve higher log-likelihood, even compared to the data used for training the generative model. 
\begin{table}[!t]
\small
\caption{\small Average log-likelihood using Parzen window for Different generative models on MNIST dataset DBN: Deep Belief Network, Stacked CAE: Stacked Contractive Auto-Encoder, Deep GSN: Deep Generative Stochastic Network \cite{bengio2014}, GAN: Generative Adversarial Network, GMMN+AE: Generative Moment Matching Network with Autoencoder, Mixture of Networks: Our model.}

\label{mnist table}
\begin{center}
\begin{sc}
\small
\begin{tabular}{lc}
\hline
Model & Average Log-Likelihood \\
\hline
\hline
DBN 	      			& 138 $\pm$ 2 	\\
Stacked CAE    			& 121 $\pm$ 1.6 \\
Deep GSN    			& 214 $\pm$ 1.1 \\
GAN				    	& 225 $\pm$ 2   \\
GMMN+AE      		    & 282 $\pm$ 2 	\\
\textbf{Mixture of Networks}   & \textbf{308} $\pm$ \textbf{2.8}\\
\hline
\end{tabular}
\end{sc}
\end{center}
\vspace{-.7cm}
\end{table}

\subsubsection{Face dataset}
The other training set we used is the Cropped Extended Yale Face Database B \cite{lee2005,georg2001}. The dataset contains $2414$ near frontal images of $38$ individuals under different illuminations. The size of each image is $32 \times 32$. We use $214$ data points for validation and the rest for training. Using autoencoder dimension is reduced from $1024$ to $128$. We employ \textit{k}-means to partition the low-dimension data into four clusters. This number is actually suggested by the knee stability method. Then, Algorithms \ref{alg:initial training} and \ref{alg:final training} are applied to the four networks, consecutively. The networks have $4$ hidden layers with $32$, $128$, $256$, and $512$ units. For this dataset, the random input is $10$-dimensional and $T_1=10$ and $T_2 = 100$. The mini batches in both steps of the algorithm contain $120$ samples.

\begin{figure}[!t]
\captionsetup[subfigure]{labelformat=empty}
    \centering
\includegraphics[trim = 00mm 0mm 00mm 00mm,width=8cm]{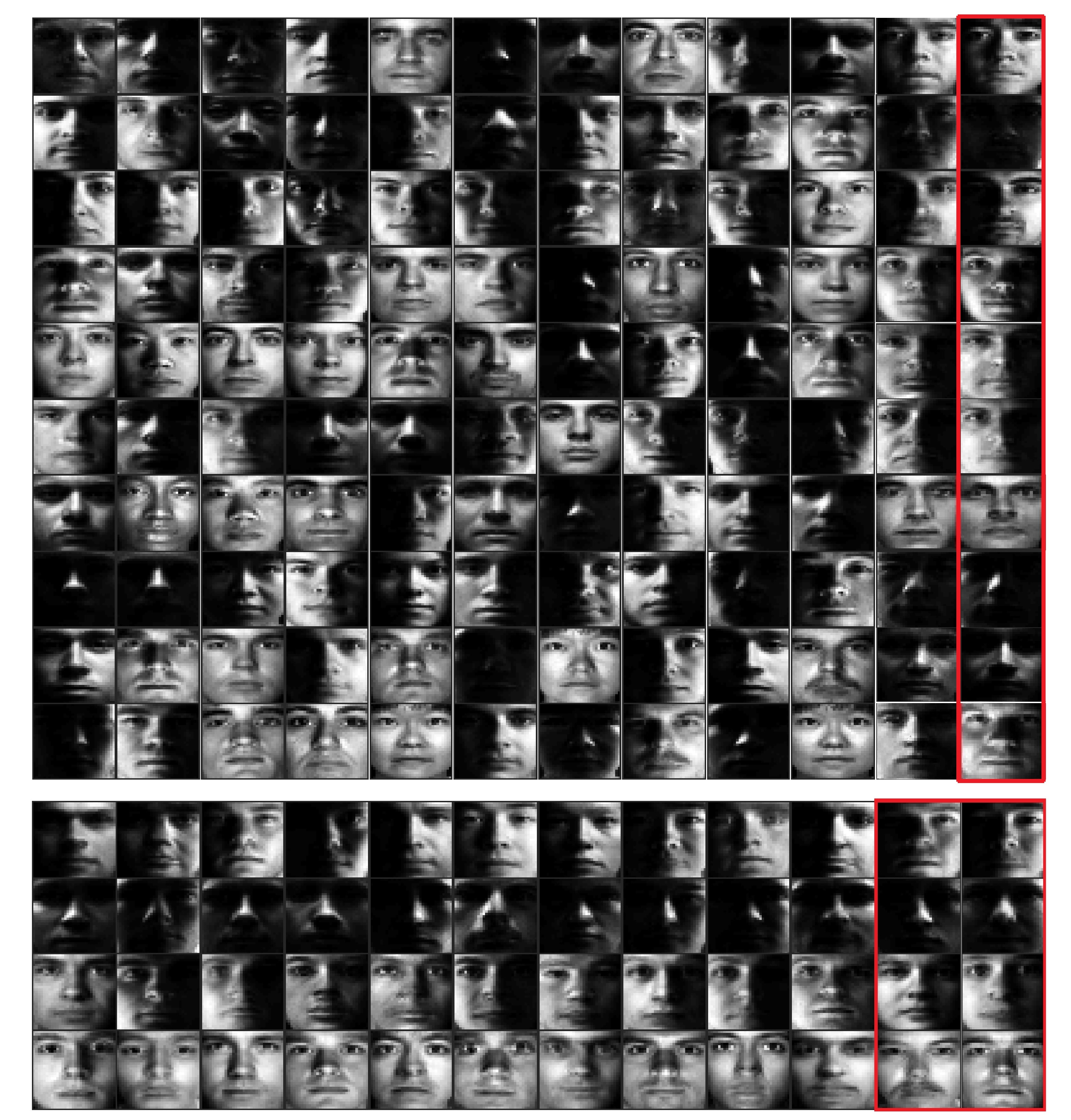} 
\caption{Top: Samples generated by the hyper-network using cluster priors. The right most column shows samples from the training set which are nearest-neighbors of their adjacent images in the low-dimension space. Bottom: Images generated by networks corresponding to each cluster. Each row is generated by one network. We can see the difference in the generated images which comes from different illuminations. The inputs to the networks for generating each of the last two columns are identical for all networks.}  
\label{fig: yale}
\end{figure}

Results of the simulations are demonstrated in Fig. \ref{fig: yale}. Networks produce images in different categories. Categories of data captured by clusters are based on lighting of the images (front lighting, sides lighting, and no lighting). We can also see the smooth changes in the faces when we pick one network and traverse in the latent space. This shows that the networks have learned a proper mapping between the latent space and the real image space.

\section{Conclusion and Future work}
We proposed an algorithm for developing a generative model using deep architectures. The algorithm has shown advantages compared to the previous generative models, which allows generating and clustering with high accuracy. The efficiency of applying the algorithm on MNIST hand-written digits and the Yale Face Database has been examined, and results support our idea. 

It will be specially interesting if a small subset of data is labeled, or when a user has clusters a small portion of data for us and we want to cluster the rest of the data accordingly. In this situation the accuracy of clustering and, consequently, the generative model will increase significantly. One application of this setting is when a hand-written text corpus is given to the model. If we label a small portion of the characters of the corpus and force the algorithm to follow the same rule for clustering the rest of the data, then we can build networks that can mimic handwriting. A related work can be found in \cite{kingma2014}. Another direction can be employing the Convolutional Neural Networks, which have shown great performance in vision tasks, instead of fully-connected networks. Then, combining the result by a natural language processing (NLP) model can be interesting. We can convert human language (text or voice) into picture, automatically.

\bibliography{Generative_Mixture_of_Networks}
\bibliographystyle{icml2016}
\end{document}